\documentclass[sigconf]{acmart}

\usepackage{float,color}
\usepackage{makecell}
\usepackage{comment} 
\usepackage[normalem]{ulem}
\usepackage{algorithmic} 
\usepackage{algorithm} 
\usepackage{array}
\usepackage{xurl}
\usepackage{subfig}
\usepackage{caption}
\usepackage{enumitem,kantlipsum}
\usepackage{placeins}
\usepackage{graphicx}

\newcommand{\gchecker}{\textsc{KompaRe}}
\newtheorem{definition}{Definition}
\newtheorem{pro-stat}{Problem Definition}
\newcommand{\hide}[1]{}
\newcommand{\hh}[1]{{\small\color{red}{\bf hh: #1}}}

\newtheorem{lemma}{Lemma}

\AtBeginDocument{%
  \providecommand\BibTeX{{%
    \normalfont B\kern-0.5em{\scshape i\kern-0.25em b}\kern-0.8em\TeX}}}

\hide{

\setcopyright{acmcopyright}
\copyrightyear{2020}
\acmYear{2020}
}

\copyrightyear{2021}
\acmYear{2021}
\setcopyright{acmlicensed}
\acmConference[Online '21]{Online '20: The 14th ACM International WSDM Conference}{ March 8-12, 2021}{Online}
\acmBooktitle{Online '21: The 14th ACM International WSDM Conference,  March 8-12, 2021}
\acmPrice{15.00}
\acmDOI{10.1145/1122445.1122456}
\acmISBN{978-1-4503-9999-9/18/06}

\begin{document}

\title{\gchecker: A Knowledge Graph Comparative Reasoning System}

\author{Lihui Liu, Boxin Du, Heng Ji, Hanghang Tong} 
\affiliation{ 
  \institution{Department of Computer Science, University of Illinois at Urbana Champaign}
}
\email{{lihuil2, boxindu2, hengji, htong}@illinois.edu }

\renewcommand{\shortauthors}{~}

\begin{abstract}

Reasoning is a fundamental capability for harnessing valuable insight, knowledge and patterns from knowledge graphs. 
Existing work has primarily been focusing on point-wise reasoning, including search, link predication, entity prediction, subgraph matching and so on.
This paper introduces {\em comparative reasoning} over knowledge graphs, which aims to infer the commonality and inconsistency 
with respect to multiple pieces of clues. We envision that the comparative reasoning will complement and expand the existing point-wise reasoning over knowledge graphs. 
In detail, we develop \gchecker, the first of its kind prototype system that provides \textit{ comparative reasoning} capability over large knowledge graphs. We present both the system architecture 
and its core algorithms, including knowledge segment extraction, pairwise reasoning and collective reasoning.
Empirical evaluations demonstrate the efficacy of the proposed \gchecker. 

\end{abstract}

\hide{

}



\maketitle


\section{Introduction}

Since its birth in 1995 ~\cite{YOO2020112965} and especially its re-introduction by Google in 2012,
knowledge graph has received more and more attentions, penetrating in a multitude of high-impact applications. To name a few, 
in fact checking, knowledge graph provides the vital background information about real-world entities and help a human fact checker corroborate or refute a claim ~\cite{Shiralkar2017}; 
in question answering, a question can be naturally formulated as a query graph, and the Q/A problem thus becomes the classic subgraph matching problem ~\cite{lei2018}; 
in recommendation systems, knowledge graph offers the auxiliary information to improve the recommendation quality and/or explainability ~\cite{Zhang2016kdd}; 
in computer vision, knowledge graph can be used to pre-optimize the model to boost its performance ~\cite{yuanijcai2017}. 
A fundamental enabling capability underlying these applications (and many more) lies in {\em reasoning}, which aims to identify errors and/or infer new conclusions from existing data ~\cite{CHEN2020112948}.
The newly discovered knowledge through reasoning provides valuable input of these down stream applications, and/or can be used to further enrich the knowledge graph itself.

Most, if not all, of the existing work on knowledge graph reasoning belongs to the {\em point-wise} approaches, which perform reasoning w.r.t. {\em a single piece of clue} (e.g., a query). For example, in knowledge graph search \cite{Wang2015QUT}, it returns the most relevant concepts for {\em a given entity}; in link prediction~\cite{SimplE}, given the `subject' and the `object' of {\em a triple}, it predicts the relation; 
in fact checking~\cite{claimbuster}, given {\em a claim} (e.g., represented as a triple of the knowledge graph), it decides whether it is authentic or falsified; in subgraph matching ~\cite{lei2018}, given {\em a query graph}, it finds exact or inexact matching subgraphs.

\begin{figure*}[ht!]
\hspace*{\fill}%
\begin{minipage}[t]{0.69\textwidth}
\centering
\vspace{0pt}
\includegraphics[width=1\textwidth, height=0.5\textwidth]{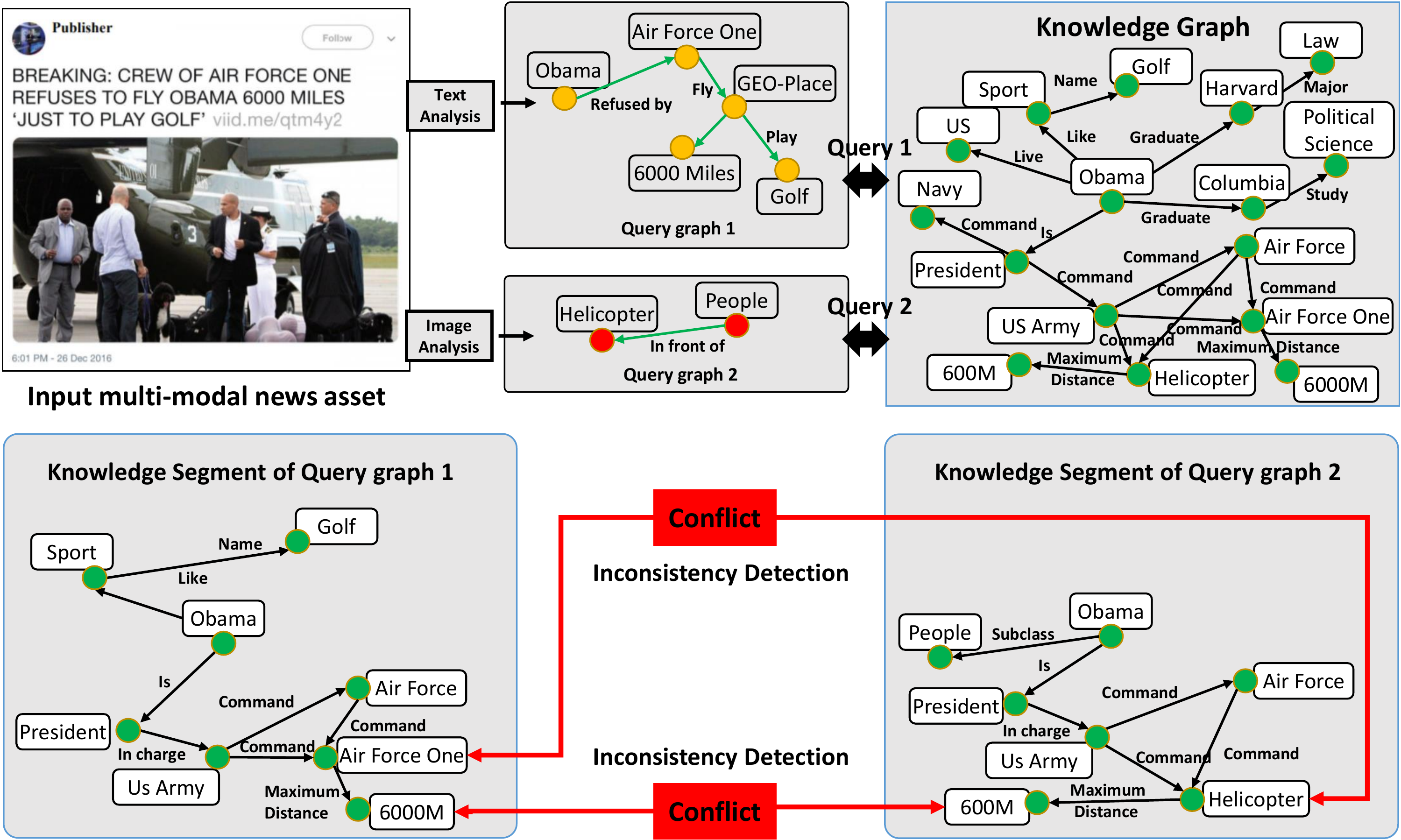}
\vspace{-2.0\baselineskip}
\caption{An illustrative example of using comparative reasoning for semantic inconsistency detection. Source of the image at the top-left: \cite{Cui2019SAMES}.
}
\label{inconsistency}
\end{minipage}%
\hspace{8.00mm}
\begin{minipage}[t]{0.26\textwidth}
\centering
\vspace{0pt}
\includegraphics[width=1\textwidth, height=1.35\textwidth]{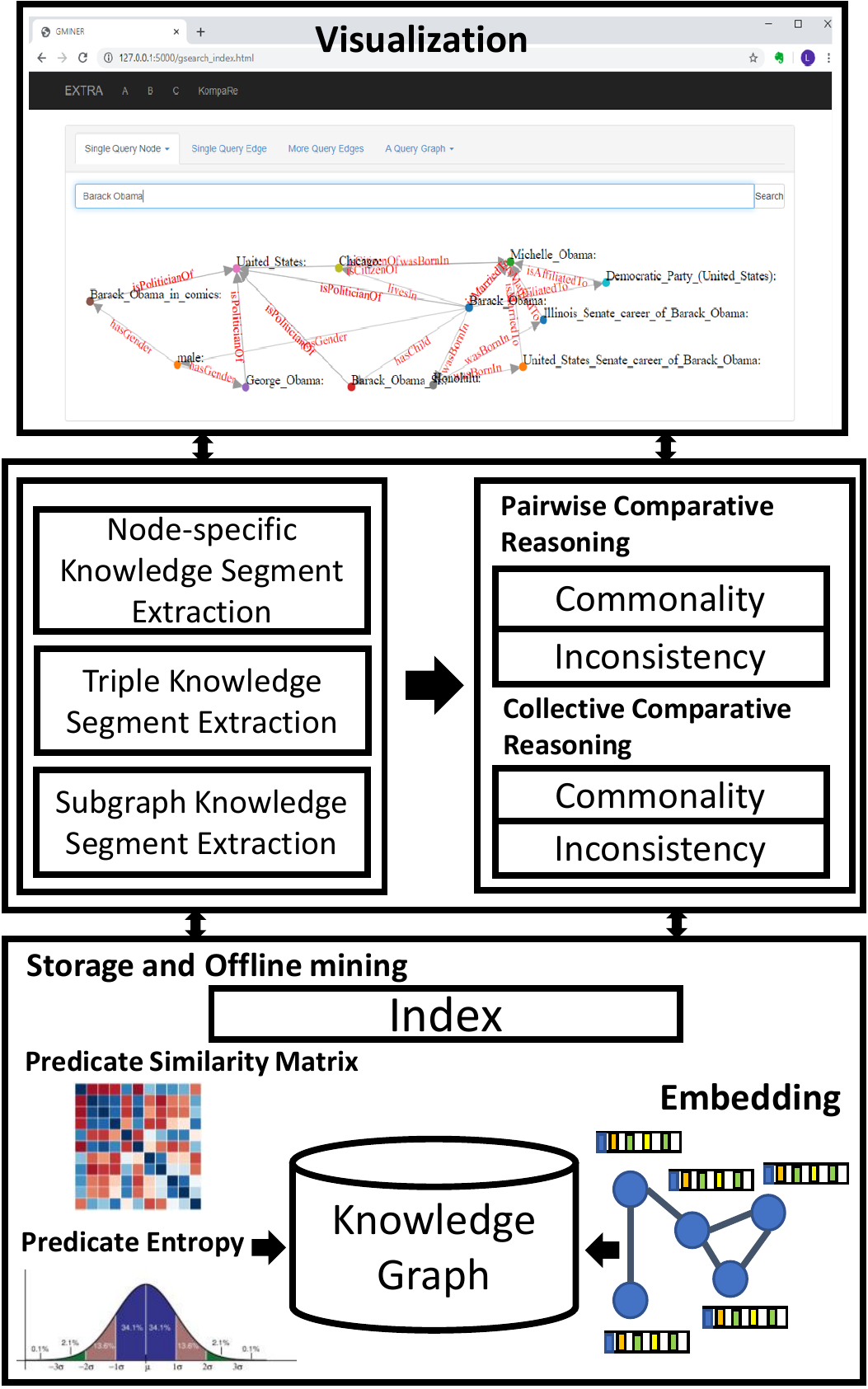}
\vspace{-2.0\baselineskip}
\caption{\gchecker\ architecture.}
\label{arch}
\end{minipage}%
\vspace{-1.0\baselineskip}
\end{figure*}

In this paper, we introduce {\em comparative reasoning} over knowledge graph, which aims to infer the commonality and/or the inconsistency with respect to multiple pieces of clues (e.g., multiple claims about a news article). We envision that the comparative reasoning will complement and expand the existing point-wise reasoning over knowledge graphs. 
This is because comparative reasoning 
offers a more complete picture w.r.t. the input clues, 
which in turn helps the users discover the subtle patterns (e.g., inconsistency) that would be invisible by point-wise approaches. 
Figure ~\ref{inconsistency} gives an example to illustrate the power of comparative reasoning. 
Suppose there is a multi-modal news asset and we wish to verify its truthfulness. To this end, two query graphs are extracted from the given news, respectively. One query graph contains all the information from the text, and the other contains the information from the image. If we perform point-wise reasoning to check each of these two query graphs {\em separately}, both seem to be true. However, if we perform reasoning w.r.t. both query graphs simultaneously, and by comparison, we could discover the subtle inconsistency between them (i.e., the different air plan types, the difference in maximum flying distances). In addition, comparative reasoning can also be used in knowledge graph expansion, integration and completion.

To be specific, we develop \gchecker, the first of its kind prototype system that provides comparative reasoning capability over large knowledge graphs. A common building block of comparative reasoning is {\em knowledge segment}, which is a small connection subgraph of a given clue (e.g., a triple or part of it) to summarize its semantic context.
Based on that, we present core algorithms to enable both {\em pairwise} reasoning and {\em collective} reasoning.
The key idea is to use influence function to discover a set of important elements in the knowledge segments.
Then, the overlapping rate and the transferred information amount 
of these important elements will help reveal commonality and inconsistency.

The main contributions of the paper are
\begin{itemize} \itemsep -1pt
    \item {\bf Problem Definition.} We introduce comparative reasoning over knowledge graphs, which complements and expands
    the existing point-wise reasoning capabilities.
    \item {\bf Prototype Systems and Algorithms.} We develop the first of its kind prototype for knowledge graph comparative reasoning, together with a suite of core enabling algorithms. 
    \item {\bf Empirical Evaluations.} We perform extensive empirical evaluations to demonstrate the efficacy of 
    \gchecker.
\end{itemize}





\vspace{-0.8\baselineskip}
\section{\gchecker\ Overview}


\begin{table*}[]
	\centering
	\small
	\caption{Summary of major functions in our system. 
	}
	\vspace{-1\baselineskip}
	\begin{tabular}{|c|c|c|c|}
		\hline
		{\bf Name} & {\bf Input}                                              & {\bf Output}         &        {\bf Key techniques}    \\ \hline
		
		\tt{f1} & A single query node                            & \makecell{A node-specific knowledge segment}  & Predicate entropy \\ \hline
		\tt{f2} & A Single query edge                                & An edge-specific knowledge segment  & \makecell{Predicate-predicate similarity} \\ \hline
		\tt{f3} & A query graph                               & \makecell{A subgraph-specific knowledge segment}  & \makecell{Semantic subgraph matching \hide{({\em Edge-Table})}} \\ \hline
		
		\tt{f4} & Two or more query edges                                & Commonality and inconsistency &  \makecell{Pairwise comparative reasoning (influence function, \\ overlapping rate, transferred information)}\\\hline
				
		\tt{f5} & A query graph                               & Commonality and inconsistency &  \makecell{Collective comparative reasoning (influence function, \\ overlapping rate, transferred information)}\\ \hline
	\end{tabular}
\label{table:fun:summary}
\vspace{-1\baselineskip}
\end{table*}


\noindent {\bf A - Architecture and Main Functions.} The architecture of \gchecker\ is shown in Figure ~\ref{arch} 
.
Generally speaking, there are three key components in  \gchecker, including (1) offline mining, (2)  online reasoning and (3) UI.

\noindent {\em (1) Offline Mining.}
{
Three are two main offline functions supported by \gchecker, including predicate entropy calculation and predicate-predicate similarity calculation\footnote{\gchecker\ also contains other offline mining algorithms, e.g. TransE ~\cite{transE}. We omit the details of these algorithms due to space limit.}.
These functions provide fundamental building blocks for \gchecker's online reasoning capabilities. 
For example, the predicate-predicate similarity will be used in both edge-specific knowledge segment extraction (Subsection~\ref{basic:edge}) and subgraph-specific knowledge segment extraction (Subsection~\ref{basic:subgraph}). 
}

\noindent {\em (2) Online Reasoning.} In the online reasoning phase, \gchecker\ supports a variety of reasoning functions which are summarized in Table ~\ref{table:fun:summary}. 
First, it supports point-wise reasoning, which returns a small connection subgraph (referred to as `knowledge segment' in this paper) for a single piece of clue provided by the user (${\tt f_1}$ to ${\tt f_3}$ in Table ~\ref{table:fun:summary}). 
For example, if the given clue is a single entity, \gchecker\ finds a semantic subgraph to summarize the context of the given entity in the underlying knowledge graph; if the given clue is a single triple, \gchecker\ finds a connection subgraph to summarize the semantic proximity from the `subject' of the triple to its `object'; 
if the given clue is a subgraph, \gchecker\ finds a semantic matching subgraph where each edge of the query graph corresponds to a knowledge segment between the two matching nodes. Second, based on these point-wise reasoning functions, \gchecker\ further supports comparative reasoning (${\tt f_4}$ and ${\tt f_5}$ in Table ~\ref{table:fun:summary}), which identifies both the commonality and the potential inconsistency w.r.t. multiple pieces of clues provided by the user. 
In addition, \gchecker\ also supports a number of common knowledge reasoning tasks, e.g., top-{\em k} query (i.e., given an entity, find the top-{\em k} most relevant entities), link prediction, subgraph matching, etc.

\noindent {\em (3) UI.} 
\gchecker\ provides a user friendly interface to visualize the point-wise and/or comparative reasoning results. 
Basically, 
the interface supports three primary functions, including (i) function selection, where the user can select different kind of functions in Table ~\ref{table:fun:summary} on the web page; (ii) query input, where the user can input various queries on the web page (e.g,. node, edge and query graph); and (iii) visualization, where 
\gchecker\ visualizes the reasoning results,and the user further modify their queries accordingly. 
The UI is implemented by HTML, Javascript and D3.js.


\noindent {\bf B - Key Challenges.} There are several challenges to implement \gchecker\ which are listed below.
First (C1 - challenge for point-wise reasoning), although there exists rich algorithms and tools to extract connection subgraphs on weighted graphs~\cite{Freitas,Tong2006CSP,Koren2006MEP}, they do not directly apply to knowledge graphs whose edges encode semantic relationship between different nodes.
Second (C2 - challenges for comparative reasoning), 
different from {\em point-wise} reasoning which focuses on a {single piece of clue}, {\em comparative reasoning} aims to infer the commonality and/or the inconsistency w.r.t. multiple clues.
Take knowledge graph based fact checking as an example. Even if each clue/claim could be true, we might still fail to detect the inconsistency between them without appropriately examining different clues/claims together.
Third (C3 - scalability), 
a common challenge to both point-wise and comparative reasoning is how to support real-time or near real-time system response over large knowledge graphs. 


\vspace{-.5\baselineskip}
\section{\gchecker\ Basics}\vspace{-.3\baselineskip}

In this section, we introduce three basic functions in our \gchecker\ system, including ${\tt f1}$, ${\tt f2}$ and ${\tt f3}$ 
in Table~\ref{table:fun:summary}.
These three functions, all of which belong to point-wise reasoning methods, form the basis of the comparative reasoning that will be introduced in the next section. 
Generally speaking, given a clue (e.g., a node, a triple or a query graph) from the user, we aim to extract a {\em knowledge segment} from the knowledge graph, which is formally defined as follows.
\vspace{-0.5\baselineskip}
\begin{definition}{\textbf{Knowledge Segment (KS for short)}} is a connection subgraph of the knowledge graph that describes the semantic context of a piece of given clue (i.e., a node, a triple or a query graph).\vspace{-0.5\baselineskip}
\end{definition}

When the given clue is a node or an edge/triple, there exist rich algorithms to extract the corresponding knowledge segment\footnote{It is worth pointing out that the extracted knowledge segment itself provides a powerful building block for several existing knowledge graph reasoning tasks, e.g. multi-hop method ~\cite{knowledge-path}, minimum cost maximum flow method ~\cite{Shiralkar2017}, etc.}
on weight graphs (e.g., a social network). To name a few,  PageRank-Nibble~\cite{nibble} is an efficient local graph partition algorithm for extracting a dense cluster w.r.t. a seed node; {\em K-}simple shortest paths based method ~\cite{Freitas} or connection subgraph  ~\cite{Faloutsos2004}, ~\cite{Koren2006MEP}
can be used to extract a concise subgraph from the source node of the querying edge to its target node. However, these methods do not directly apply to knowledge graphs because the edges (i.e., predicates) of a knowledge graph have specific semantic meanings (i.e., types, relations). To address this issue, we seek to convert the knowledge graph to a weighted graph by designing (1) a predicate entropy measure for node-specific knowledge segment extraction (Subsection~\ref{basic:node}), 
and (2) a predicate-predicate similarity measure for edge-specific knowledge segment extraction (Subsection~\ref{basic:edge}), respectively. 

When the given clue itself is a subgraph (Subsection~\ref{basic:subgraph}), we propose to extract a {\em semantic matching subgraph}. 
We would like to point out that semantic matching subgraph extraction is similar to but bears subtle difference from the traditional subgraph matching problem ~\cite{lihui}. 
In subgraph matching, it aims to find a matching edge or path for each pair of matching nodes if they are required to be connected by the query graph; whereas in semantic subgraph matching, we aim to find a small connection subgraph (i.e., an edge-specific knowledge segment) for each pair of matching nodes that are required to be connected according to the query subgraph. In other words, a subgraph-specific knowledge segment consists of multiple inter-linked edge-specific knowledge segments (i.e., one edge-specific knowledge segment for each edge of the input query subgraph). 
We envision that the subgraph-specific knowledge segment provides richer semantics, including both the semantics for each edge of the query graph and the semantics for the relationship between different edges of the input query graph. 

\vspace{-1\baselineskip}
\subsection{Node-specific Knowledge Segment}~\label{basic:node}
PageRank-Nibble ~\cite{nibble} is a local graph partitioning algorithm to find a dense cluster near a seed node (i.e., the query node) on a weighted graph. 
It calculates the approximate PageRank vector with running time independent of the graph size. 
By sweeping over the PageRank vector, it finds a cut with a small conductance to obtain the local partition. In order to apply PageRank-Nibble to find node-specific knowledge segment, we propose to convert the knowledge graph into a weighted graph by {\em predicate entropy}. 

To be specific, we treat each predicate in the knowledge graph as a random variable. The entropy of the predicates offers a natural way to measure its uncertainty and thus can be used to quantify the importance of the corresponding predicate. For example, some predicates have a high degree of uncertainty, e.g. {\tt livesIn}, {\tt isLocatedIn}, {\tt hasNeighbor}, {\tt actedIn}. 
This is because, in knowledge graph, different persons usually have different numbers of neighbors, and different actors may act in different movies. 
A predicate with high uncertainty indicates that it is quite common which offers little specific semantics of the related entity, and thus it should have low importance.
On the other hand, some predicates have a low degree of uncertainty, e.g. {\tt isPresident}, 
{\tt isMarriedTo}. 
This is because only one person can be the president of a given country at a time, and for most of persons, they marry once in life. 
Such a predicate often provides very specific semantics about the corresponding entity and thus it should have high importance. Based on this observation, we propose to use predicate entropy to measure the predicate importance as follows.

We treat each entity and all the predicates surrounding it as the outcome of an experiment. In this way, we could obtain different distributions for different predicates. 
Let $i$ denote a predicate in the knowledge graph, and $D$ denote the maximal out-degree of a node. For a given node, assume it contains $d$ out links whose label is $i$, we have $0 \le d \le D$. Let $\mathcal{V}_i^d$ denote the node set which contains $d$ out links with label $i$, $\textrm{E}$ denote the entropy, and
$\mathbf{P}_{i}^d$ denote the probability of a node having $d$ out links with label/predicate $i$.
The entropy of a given predicate $i$ can be computed as $ \textrm{E}(i) = \sum_{d=1}^{D} -\mathbf{P}_{i}^d \log(\mathbf{P}_{i}^d)$, where $ 
    \mathbf{P}_{i}^d = \frac{|\mathcal{V}_i^d|}{\sum_{d=1}^D|\mathcal{V}_i^d|}$. Finally, we compute the importance of a predicate $i$ as $w(i) = 2\sigma(\frac{1}{\textrm {E}(i)}) - 1$, where $\sigma()$ is the sigmoid function.


\subsection{Edge-specific Knowledge Segment}~\label{basic:edge}
Edge-specific knowledge segment extraction aims at finding a knowledge segment to best characterize the semantic context of the given edge (i.e. a triple). Several connection subgraph extraction methods exist for a weighted graph, e.g. ~\cite{Tong2006CSP}, ~\cite{Koren2006MEP}, ~\cite{Freitas}. 
We propose to use a TF-IDF based method\footnote{The TF-IDF based method was also used in~\cite{Shiralkar2017} for computational fact checking.} to measure the similarity between different predicates, and transfer the knowledge graph into a weighted graph whose edge weight represents the similarity between the edge predicate and query predicate. Then, we find {\em k-}simple shortest paths~\cite{Koren2006MEP} from the subject to the object of the given query edge as its knowledge segment. 

The key idea behind predicate similarity is to treat each triple in the knowledge graph and its adjacent neighboring triples as a document, and use a TF-IDF like weighting strategy to calculate the predicate similarity. 
Consider a triple $e_t$ = <${\tt s}$, {\tt receiveDegreeFrom}, ${\tt o}$> in the knowledge graph whose predicate is $i = {\tt receiveDegreeFrom}$. In the neighborhood of $e_t$,
there is a high probability that triples like <{\tt s}, {\tt major}, {\tt o}> and <{\tt s}, {\tt graduateFrom}, {\tt o}> 
also exist (adjacent to $e_t$). 
The predicates of these triples should have high similarity with each other. On the other hand, triples like <{\tt s}, {\tt livesIn}, {\tt o}>, <{\tt s}, {\tt hasNeighbor}, {\tt o}> may also occur in the adjacent neighborhood of triple $e_t$. This is because these predicates 
are very common in the knowledge graph, and occur almost everywhere. These predicates are like the stop words such as “the”, “a”, “an” in a document. Therefore, if we treat each predicate and its neighborhood as a document, we could use a TF-IDF like weighting strategy to find highly similar predicates and in the meanwhile penalize common predicates like {\tt livesIn}, {\tt hasNeighbor}.

To be specific, we use the knowledge graph to build a co-occurrence matrix of predicates, and calculate their similarity by a TF-IDF like weighting strategy as follows. Let $i,j$ denote two different predicates. We define the $\textrm{TF}$ between two predicates as $\textrm{TF}(i, j) = \log(1 + C(i,j) {w}(j))$, where $C(i,j)$ is the co-occurrence of predicate $i$ and $j$. The $\textrm{IDF}$ is defined as $\textrm{IDF}(j) = \log \frac{|M|}{|\{i : C(i,j)>0\}|}$, 
where $M$ is the number of predicates in the knowledge graph. Then, we build a TF-IDF weighted co-occurrence matrix $U$ as $U(i, j) = \textrm{TF}(i, j) \times IDF(j)$.
Finally, the similarity of two predicates is defined as $\textrm{Sim(i, j)} = \textrm{Cosine}(U_i, U_j)$, where
where $U_i$ and $U_j$ are the $i^{th}$ row and $j^{th}$ row of $U$, respectively.

\subsection{Subgraph-specific Knowledge Segment}~\label{basic:subgraph}
\hide{
\begin{figure}
	\centering
	\includegraphics[width=0.5\textwidth,height=0.2\textwidth]{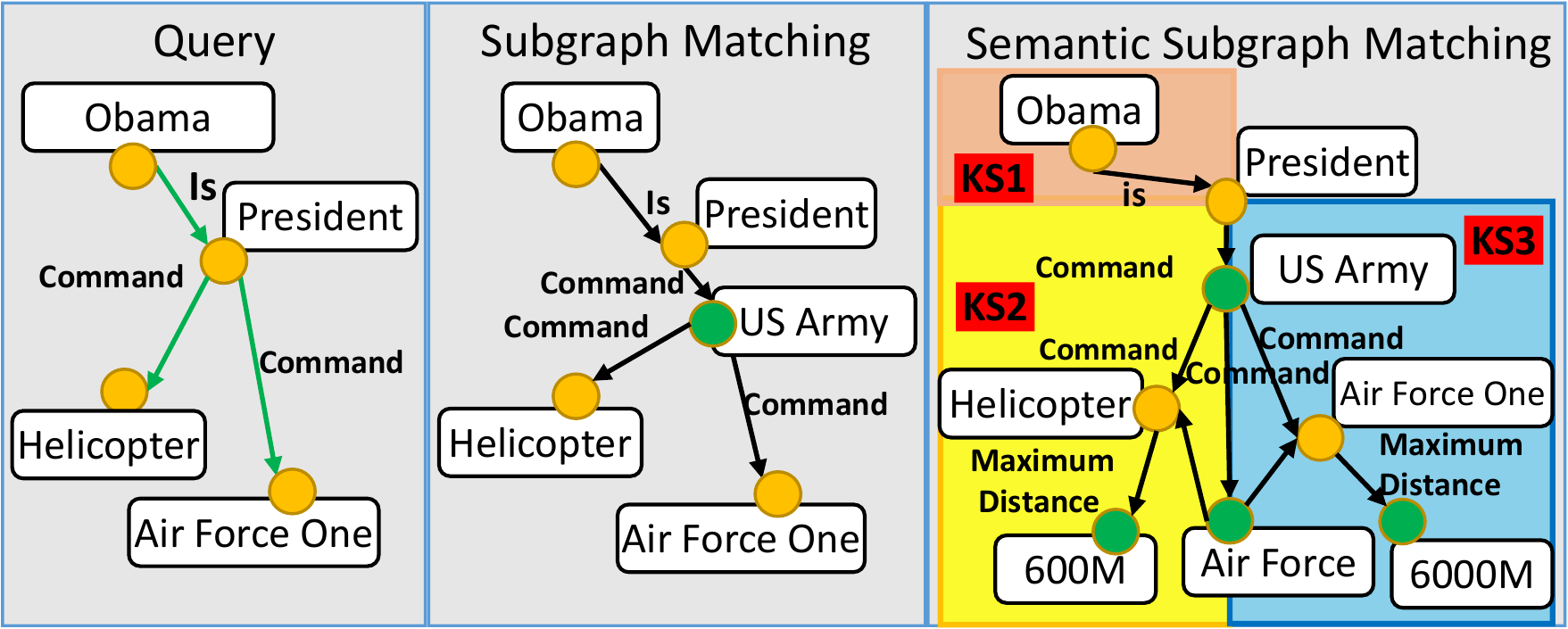}
	\vspace{-2\baselineskip}
	\caption{Difference between subgraph matching and semantic subgraph matching. For semantic subgraph matching, different color show different knowledge segments.}
	\label{sub-sem-diff}
	\vspace{-2\baselineskip}
\end{figure}
}

Given an attributed query graph $\mathcal{Q}$=\{$V^Q$, $E^Q$, $L^Q$\},
the traditional subgraph matching aims to find an edge or a path for each $e_i \in E^Q$. On the contrary, 
subgraph-specific knowledge segment extraction aims to find an edge-specific knowledge segment for each edge $e_i \in E^Q$. 
To our best knowledge, there is no existing method for subgraph-specific knowledge segment extraction. 
\hide{
We generalize a recent indexing-based subgraph matching algorithm called GFinder~\cite{lihui}. We design a new data structure called {\em Edge-Table} to index the knowledge segment between each pair of matching nodes to reduce memory cost. 
For each edge $e_i \in E^Q$, there is a corresponding {\em Edge-Table} to store the information of its corresponding knowledge segment, e.g. 
triples in the knowledge segments, the {\em k-}simple shortest paths and their costs. 
}
In order to find the edge-specific knowledge segments for each $e_i \in E^Q$, we again use the {\em k-}simple shortest path method to extract the paths with the lowest cost. The cost of a path is equal to the sum of the reciprocal of the predicate-predicate similarity of all edges in the path. Finally, all the edge-specific knowledge segments will be merged together to obtain the semantic matching subgraph (i.e., the subgraph-specific knowledge segment).

\section{\gchecker\ Comparative Reasoning}

In this section, we introduce the technical details of comparative reasoning in \gchecker. We first introduce the pairwise reasoning ({\tt f4} in Table~\ref{table:fun:summary}) for two pieces of clues (e.g., two edges/triples), and then present the collective comparative reasoning ({\tt f5} in Table~\ref{table:fun:summary}). Table ~\ref{notation} summarizes the main notation used in this section.

\vspace{-0.6\baselineskip}
\subsection{Pairwise Comparative Reasoning}

Pairwise comparative reasoning aims to infer the commonality and/or inconsistency with respect to a pair of clues according to their knowledge segments. Here, we assume that the two given clues are two edges/triples: $E^Q_1 = <{\tt s_1}, {\tt p_1}, {\tt o_1}>$ and $E^Q_2 = <{\tt s_2}, {\tt p_2}, {\tt o_2}>$ where ${\tt s_1}, {\tt o_1}, {\tt s_2}, {\tt o_2} \in V^Q$ and ${\tt p_1}, {\tt p_2} \in E^Q$. We denote their corresponding knowledge segments as $KS_1$ for $E^Q_1$ and $KS_2$ for $E^Q_2$, respectively. 
The commonality and inconsistency between these two knowledge segments are defined as follows. 
\vspace{-0.6\baselineskip}
\begin{definition}{\textbf{Commonality.}}
Given two triples ($E^Q_1$ and $E^Q_2$) and their knowledge segments ($KS_1$ and $KS_2$), the commonality of these two triples refers to the shared nodes and edges between $E^Q_1$ and $E^Q_2$, as well as the shared nodes and edges between $KS_1$ and $KS_2$: $((V^{KS_1} \cap V^{KS_2}) \cup (V^{Q_1} \cap V^{Q_2}) , (E^{KS_1} \cap E^{KS_2}) \cup (E^{Q_1} \cap E^{Q_2})  )$.
\end{definition}
\vspace{-0.8\baselineskip}
\begin{definition}{\textbf{Inconsistency.}}
Given two knowledge segments $KS_1$ and $KS_2$, 
the inconsistency between these two knowledge segments refers to any element (node, node attribute or edge) in $KS_1$ and $KS_2$ that contradicts with each other. 
\end{definition}
\vspace{-0.5\baselineskip}

\begin{table}[]
	\centering
	\footnotesize
	\caption{Notations and definition}
	\vspace{-1.5\baselineskip}
	\begin{tabular}{|c|l|}
		\hline
		{\bf Symbols}       & {\bf Definition}                \\ \hline
		$\mathcal{Q}$=\{$V^Q$, $E^Q$, $L^Q$\} & an attributed query graph \\ \hline
		$\mathcal{G}$=\{$V^G$, $E^G$, $L^G$\} & a knowledge graph  \\ \hline
		$KS_i$         & knowledge segment $i$ \\ \hline
		$A_i$         & adjacency matrix of $KS_i$ \\ \hline
		$N_i$         & {attribute matrix of $KS_i$, the $j^\textrm{th}$ row} \\  
		 ~ & {denotes the attribute vector of node $j$ in $KS_i$} \\ \hline
		$A_{\times}$      & Kronecker product of $A_1$ and $A_2$      \\ \hline
		$N^l$      & diagonal matrix of the $l^\textrm{th}$ node attribute  \\ \hline
		$N_{\times}$      &  combined node attribute matrix  \\ \hline
		$S^{i,j}$          & single entry matrix $S^{i,j}(i,j)=1$ and zeros elsewhere  \\ \hline
	\end{tabular}
\label{notation}
\vspace{-1.5\baselineskip}
\end{table}

In order to find out if the two given edges/triples are inconsistent, we first need to determine if they refer to the same thing/fact. Given a pair of clues $<{\tt s_1}, {\tt p_1}, {\tt o_1}>$ and $<{\tt s_2}, {\tt p_2}, {\tt o_2}>$, we divide it into the following six cases. including
\noindent 
\begin{enumerate}[wide, labelwidth=!, labelindent=0pt]
\item [C1.]
{
${\tt s_1} \neq {\tt s_2}$, ${\tt s_1} \neq {\tt o_2}$, ${\tt o_1} \neq {\tt s_2}$, ${\tt o_1} \neq {\tt o_2}$. 
For this case, these two clues apparently refer to different things,
e.g., <{\tt Barack Obama}, {\tt wasBornIn}, {\tt Honolulu}> and <{\tt Google}, {\tt isLocatedIn}, {\tt USA}>.
}
\item [C2.]
{
${\tt s_1} = {\tt s_2}$ and ${\tt o_1} = {\tt o_2}$. 
If ${\tt p_1} = {\tt p_2}$, these two clues are equal. 
If ${\tt p_1}$ and ${\tt p_2}$ are different or irrelevant, e.g., ${\tt p_1} ={\tt wasBornIn}$, ${\tt p_2} = {\tt hasWebsite}$, 
these two clues refer to different things.
However, if ${\tt p_1}$ is opposite of ${\tt p_2}$, they are inconsistent with each other. 
}
\item [C3.]
{
${\tt s_1} = {\tt s_2}$ but ${\tt p_1} \neq {\tt p_2}$ and ${\tt o_1} \neq {\tt o_2}$, e.g., <{\tt Barack Obama}, {\tt wasBornIn}, {\tt Honolulu}>, <{\tt Barack Obama}, {\tt graduatedFrom}, {\tt Harvard}>. 
}
\item [C4.]
{
${\tt s_1} = {\tt s_2}$, ${\tt p_1} = {\tt p_2}$, but ${\tt o_1} \neq {\tt o_2}$, 
e.g., <{\tt Barack Obama}, {\tt wasBornIn}, {\tt Honolulu}>, <{\tt Barack Obama}, {\tt wasBornIn}, {\tt Hawaii}>. 
}
\item [C5.]
{
${\tt o_1} = {\tt o_2}$, but ${\tt s_1} \neq {\tt s_2}$. 
For this case, no matter what ${\tt p_1}$ and ${\tt p_2}$ are, these two clues refer to different things.
}
\item [C6.]
{
${\tt o_1} = {\tt s_2}$. 
For this case, no matter what ${\tt p_1}$ and ${\tt p_2}$ are, they refer to different things.
For example, <{\tt Barack Obama}, {\tt is}, {\tt President}>, <{\tt President}, {\tt workAt}, {\tt White House}>.
}
\end{enumerate}

Among these six cases, we can see that the clue pair in C1, C5 and C6 refer to different things. Therefore, there is no need to check the inconsistency between them. For C2, we only need to check the semantic meaning of their predicates, i.e., if  ${\tt p_1}$ is the opposite of ${\tt p_2}$. For example, ${\tt p_1} = {\tt isFather}$ and ${\tt p_2} = {\tt isSon}$, they are inconsistent with each other. Otherwise, there is no inconsistency between them.
We mainly focus on C3 and C4 where the two clues may be inconsistent with each other even if each of them is true. For example, in Figure ~\ref{inconsistency}, either <{\tt Barack Obama}, {\tt refusedBy}, {\tt Air Force One}> or <{\tt Barack Obama}, {\tt inFront} , {\tt Helicopter}> could be true. But putting them together, they cannot be both true, since these two claims could not happen at the same time. In other words, they are mutually exclusive with each other and thus are inconsistent. 
However,
queries like <{\tt Barack Obama}, {\tt wasBornIn}, {\tt Honolulu}> and <{\tt Barack Obama}, {\tt wasBornIn}, {\tt Hawaii}> are both true, because {\tt Honolulu} belongs to {\tt Hawaii}. Alternatively, we can say that {\tt Hawaii} contains {\tt Honolulu}. 
Another example is <{\tt Barack Obama}, {\tt wasBornIn}, {\tt Honolulu}> and <{\tt Barack Obama}, {\tt is}, {\tt President}>, both of which are true. Although they have the same subject, they refer to two different things.
We summarize that if (1) the subjects of two clues are the same, and (2) their predicates are similar with each other or the same, they refer to the same thing. Furthermore, if their objects are two uncorrelated entities, it is high likely that these two clues are inconsistent with each other. 

Based on the above observations, we take the following three steps for pairwise comparative reasoning. First, given a pair of clues, we decide which of six cases it belongs to, by checking the subjects, predicates and objects of these two clues. Second, if this pair of clues belongs to C3 or C4, we need to decide whether they refer to the same thing or two different things. 
To this end, we first find a set of key elements (nodes or edges or node attributes) in these two knowledge segments. If most of these key elements belong to the commonality of these two knowledge segments, it is high likely that they refer to the same thing. Otherwise, these two clues refer to different things. Third, if they refer to the same thing, we further decide whether they conflict with each other. Here, the key idea is as follows. We build two new query triples <${\tt o_1}$, {\tt isTypeOf}, ${\tt o_2}$> and <${\tt o_2}$, {\tt isTypeOf}, ${\tt o_1}$>. If one of them is true, the original two triples are consistent with each other. Otherwise, they are inconsistent.





In order to find the key elements, we propose to use the influence function w.r.t. the knowledge segment similarity \cite{qinghai}. The basic idea is that if we perturb a key element (e.g., change the attribute of a node or remove a node/edge), it would have a significant impact on the overall similarity between these two knowledge segments. Let $KS_1$ and $KS_2$ be the two knowledge segments. 
We can treat the knowledge segment as an attributed graph, where different entities have different attributes. 
We use random walk graph kernel with node attribute to measure the similarity between these two knowledge segments ~\cite{qinghai}. 

\vspace{-1\baselineskip}
\begin{eqnarray}\label{eq:rwgraphkernel}
\textrm{Sim}(KS_1, KS_2) = q'_{\times}(I - cN_{\times}A_{\times})^{-1}N_{\times}P_{\times}
\end{eqnarray}
where ${q'}_{\times}$ and $p_{\times}$ are the stopping probability distribution and the initial probability distribution of random walks on the product matrix, respectively. $N_{\times}$ is the combined node attribute matrix of the two knowledge segments $N_{\times} = \sum_{j=1}^d N_1^j \otimes N_2^j$ where $N_i^j$ ($i\in \{1,2 \}$) is the diagonal matrix of the $j^\textrm{th}$ column of attribute matrix $N_i$. 
$A_{\times}$ is the Kronecker product of the adjacency matrices of knowledge segments $A_1$ and $A_2$. $0 < c < 1$ is a parameter.

We propose to use the influence function of  $\textrm{Sim}(KS_1 , KS_2)$ w.r.t. knowledge segment elements 
 $\frac{\partial{Sim(KS_1, KS_2)}}{\partial {e}}$,
where $e$ represents an element of the knowledge segment $KS_1$ or $KS_2$.
The element with a high absolute influence function value is treated as a key element, and it can be a node, an edge, or a node attribute. Specifically, we consider three kinds of influence functions w.r.t. the elements in $KS_1$, including edge influence, node influence and node attribute influence, which can be computed according to the following lemma. Note that the influence function w.r.t. elements in $KS_2$ can be computed in a similar way, and thus is omitted for space. 
\vspace{-0.3\baselineskip}
\begin{lemma}\label{lm:influence}\noindent (Knowledge Segment Similarity Influence Function ~\cite{qinghai}.)
Given $\textrm{Sim}(KS_1, KS_2)$ in Eq.~\eqref{eq:rwgraphkernel}. Let $Q = (I - c N_{\times} A_{\times})^{-1}$ and $S^{j,i}$ is a single entry matrix defined in Table~\ref{notation}. We have that
\item [] (1.) The influence of an edge $A_1(i,j)$ in $KS_1$ can be calculated as 
$I(A_1(i,j))= \frac{\partial \textrm{Sim}(KS_1, KS_2)}{\partial A_1(i,j)} = 
c {q'}_{\times}QN_{\times}[(S^{i,j} + S^{j,i})\otimes A_2]QN_{\times}p_{\times}$. 

\item [] (2.) The influence of a node $i$ in $KS_1$ can be calculated as 
$I(N_1(i)) = \frac{\partial \textrm{Sim}(KS_1, KS_2)}{\partial N_1(i)} = 
c {q'}_{\times}QN_{\times}[\sum_{j|A_1(i,j)=1}(S^{i,j} + S^{j,i})\otimes A_2]QN_{\times}p_{\times}$. 

\item [] (3.) The influence of a node attribute $j$ of node $i$ in $KS_1$ can be calculated as 
$I(N_1^j(i, i)) = \frac{\partial \textrm{Sim}(KS_1, KS_2)}{\partial N_1^j(i, i)} = 
q'_{\times}Q[S^{i,i}\otimes N_2^j](I + c A_{\times}QN_{\times})p_{\times}$. 

\end{lemma}
\vspace{-0.3\baselineskip}
Note that according to Lemma~\ref{lm:influence}, if an element only belongs to $KS_1$ or $KS_2$, its influence function value will be $0$. In order to avoid this, we introduce a fully connected background graph to $KS_1$ and $KS_2$, respectively. 
This background graph contains all the nodes in $KS_1$ and $KS_2$, and it is disconnected with $KS_1$ and $KS_2$. If we treat $KS_1$ and $KS_2$ as two documents, we can think of this background graph as the background word distribution in a language model.

For a given knowledge segment, we flag the top 50\% of the elements (e.g., node attribute, node and edge) with the highest absolute influence function values as key elements. 
We would like to check whether these key elements belong to the commonality of the two knowledge segments. If most of them (e.g. 60\% or more) belong to the commonality of these two knowledge segments, we say the two query clues describe the same thing. Otherwise, they refer to different things and thus we do not need to check the inconsistency between them. 

If we determine that the query clues refer to the same thing, the next step is to decide whether they are inconsistent with each other. That is, given query clues <${\tt s_1}$, ${\tt p_1}$, ${\tt o_1}$> and <${\tt s_1}$, ${\tt p_2}$, ${\tt o_2}$>, we need to decide whether ${\tt o_1}$ belongs to ${\tt o_2}$ or ${\tt o_2}$ belongs to ${\tt o_1}$. 
To this end, we build two new queries <${\tt o_1}$, {\tt isTypeOf}, ${\tt o_2}$> and <${\tt o_2}$, {\tt isTypeOf}, ${\tt o_1}$>. Then, we extract the knowledge segments for these two queries, and check whether these two segments are true. If one of them is true, we say the original clues are consistent with each other, otherwise they are inconsistent. 
After we extract the knowledge segments for <${\tt o_1}$, {\tt isTypeOf}, ${\tt o_2}$> and <${\tt o_2}$, {\tt isTypeOf}, ${\tt o_1}$>, we treat each knowledge segment as a directed graph, and calculate how much information can be transferred from the subject to the object. We define the transferred information amount 
as:
\vspace{-0.3\baselineskip}
\begin{equation}\label{eq:transinfo}
    \textrm{infTrans}({\tt o_1}, {\tt o_2}) = \max_{1 \leq j \leq k} \textrm{pathValue}(j)
    \vspace{-0.3\baselineskip}
\end{equation} where $\textrm{pathValue}(j)$ is defined as the multiplication of the weights in the path. For an edge, its weight is the predicate-predicate similarity $\textrm{Sim}({\tt isTypeOf}, e_i)$. If $\max\{\textrm{infTrans}({\tt o_1}, {\tt o_2}),  \textrm{infTrans}({\tt o_2}, {\tt o_1})\}$ is larger than a threshold $T$, then we say ${\tt o_1}$ belongs to ${\tt o_2}$ or ${\tt o_2}$ belongs to ${\tt o_1}$. We set $T = 0.700$ in our experiment.

\hide{
\begin{figure}
	\centering
	\includegraphics[width=0.43\textwidth]{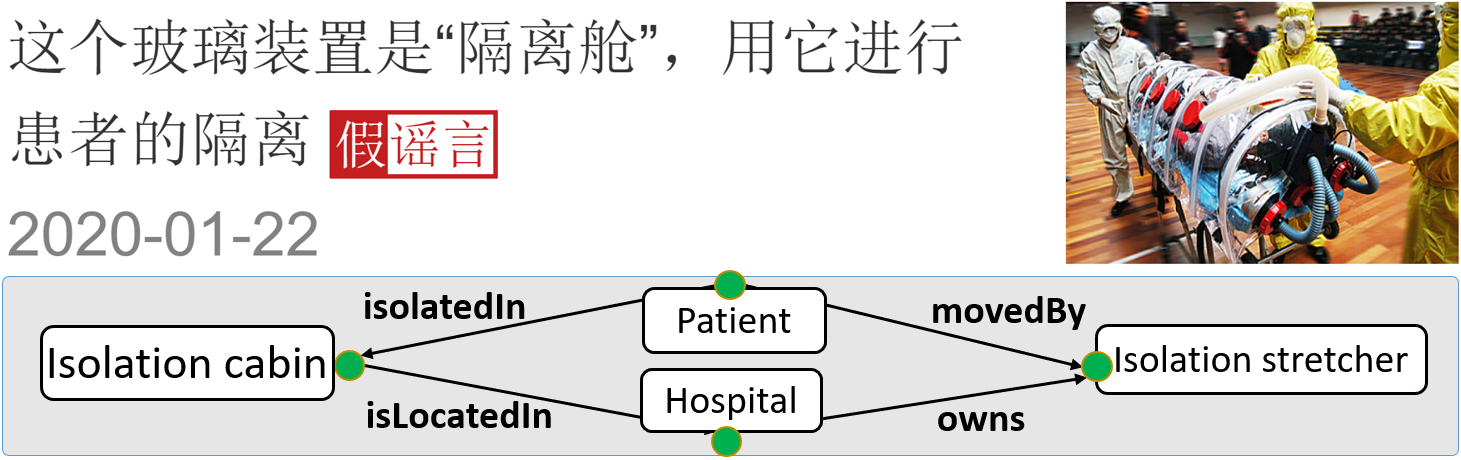}
	\vspace{-1\baselineskip}
	\caption{A fake news example from Wuhan coronavirus crisis. The text in the news says ``{\em This glass device is `Isolation Cabin', which is used to  isolate patients.}".}
	\label{qq_example}
	\vspace{-1\baselineskip}
\end{figure}

Let us take an example from Wuhan coronavirus crisis which is shown in Figure ~\ref{qq_example}. 
The query we get from text is <{\tt Patient}, {\tt isolatedIn}, {\tt Isolation Cabin}>,
and the query we get from image is <{\tt Patient}, {\tt isolatedIn}, {\tt Isolation stretcher}>.
To detect the inconsistency of this pair of clues, we only need to check whether <{\tt Isolation Cabin}, {\tt isTypeOf}, {\tt Isolation stretcher}> or <{\tt Isolation stretcher}, {\tt isTypeOf}, {\tt Isolation Cabin}>.
If we have $\textrm{infTrans}({\tt Isolation~Cabin}, {\tt Isolation~stretcher}) < 0.700$ based on the knowledge segment extracted from the knowledge graph, we conclude that Isolation Cabin is not Isolation stretcher. Therefore, there exists inconsistency in this news.
}


\vspace{-0.8\baselineskip}
\subsection{Collective Comparative Reasoning}

Different from pairwise comparative reasoning, collective comparative reasoning aims to find the commonality and/or inconsistency inside a query graph which consists of a set of inter-connected edges/triples. We first give the corresponding definition below.

\begin{definition}{\textbf{Collective Commonality.}}
For each edge $E^Q_i$ in a query graph $Q$, let $KS_i$ be its knowledge segment. 
The collective commonality between any triple pair in the query graph is the intersection of their knowledge segments.
\end{definition}

\begin{definition}{\textbf{Collective Inconsistency.}}
For each edge $E^Q_i$ in a query graph $Q$, let $KS_i$ be its knowledge segment. 
The collective inconsistency refers to any elements (node or edge or node attribute) in these knowledge segments that contradict with each other.
\end{definition}

To check the inconsistency, one naive method is using the pairwise comparative reasoning method to check the inconsistency for each pair of edges in the query graph. However, this method is neither computationally efficient nor sufficient. For the former, if two clues (e.g., two claims from a news article) are weakly or not related with each other on the query graph, we might not need to check the inconsistency between them at all. For the latter, in some subtle situation, the semantic inconsistencies could {\em only} be identified when we collectively reason over multiple (more than two) knowledge segments. For example, given the following three claims, including 
(1) {\em Obama is refused by Air Force One}; 
(2) {\em Obama is the president of the US};
(3) {\em The president of US is in front of a helicopter}.
Only if we reason these three claims collectively, can we identify the semantic inconsistency among them. 

\begin{figure}
	\centering
	\vspace{-1\baselineskip}
	\includegraphics[width=0.42\textwidth, height=0.2\textwidth]{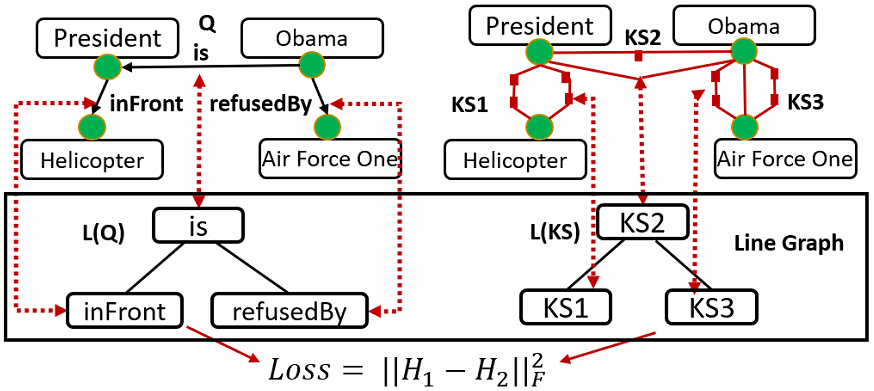}
	\vspace{-1\baselineskip}
	\caption{Collective comparative reasoning workflow.}
	\label{collective-compare-workflow}
	\vspace{-1\baselineskip}
\end{figure}

Based on the above observation, we propose the following method to detect the collective inconsistency.
\textbf{First}, we find a set of key elements inside the semantic matching subgraph. 
Different from pair-wise comparative reasoning, the importance/influence of an element for collective comparative reasoning is calculated by the entire semantic matching subgraph.
More specifically, 
we first transform the query graph and its semantic matching subgraph (i.e., subgraph-specific knowledge segment) into two line graphs, which are defined as follows.
\vspace{-0.5\baselineskip}
\begin{definition} {\bf Line Graph~\cite{Shiralkar2017}}.
For an arbitrary graph $G=(V, E)$, the line graph $L(G) = (V', E')$ of  $G$ has the following properties: 
(1) the node set of $L(G)$ is the edge set of $G$ ($V' = E$);
(2) two nodes $V'_i$, $V'_j$ in $L(G)$ are adjacent if and only if the corresponding edges $e_i$, $e_j$ of $G$ are incident on the same node in $G$.
\end{definition}
\vspace{-0.5\baselineskip}
Figure ~\ref{collective-compare-workflow} gives an example of the line graph. 
For the line graph $L(Q)$, the edge weight is the predicate-predicate similarity of the two nodes it connects.
For the line graph $L(KS)$, the edge weight is the knowledge segment similarity by Eq.~\eqref{eq:rwgraphkernel} of the two nodes it connects. 
The rationality of building these two line graphs is that if the semantic matching subgraph is a good representation of the original query graph, the edge-edge similarity in $L(Q)$ would be similar to the knowledge segment similarity in $L(KS)$

To measure the importance of an element, we propose to use the influence function w.r.t. the distance between $L(Q)$ and $L(KS)$. We assume that a key element, if perturbed, would have a great effect on the distance
$\textrm{Loss} = || H_1 - H_2 ||_F^2$ 
, where $H_1$ is the weighted adjacency matrix of $L(Q)$, and $H_2$ is the weighted adjacency matrix of $L(KS)$. We use the influence function $\frac{\partial \textrm{Loss}(H_1, H_2)}{\partial e}$, where $e$ represents an element of the knowledge segment graph and it could be a node, an edge, or a node attribute. Lemma~\ref{lm:collectiveinfluence} provides the details on how to compute such influence functions. We skip the proof of Lemma ~\ref{lm:collectiveinfluence} since it is very easy.


\begin{lemma}\label{lm:collectiveinfluence}
Given the loss function $Loss = || H_1 - H_2 ||_F^2$. Let $n$, $k$ denote two different nodes in $L(Q)$,
and $KS_n$, $KS_k$ denote their corresponding knowledge segments. 
Let $h_{e_{k,n}}$ denote the weight of edge between node $k$ and $n$,  and $h_{c_{k,n}}$ denote the weight of edge between $KS_k$ and $KS_n$. We have

\item [] (1.) The influence of an edge $A_n(i,j)$ in knowledge segment $KS_n$ can be calculated as $I(A_n(i,j)) = \sum_{k \in N(n)} -2(h_{e_{k,n}} - h_{c_{k,n}})\frac{\partial sim(KS_n, KS_k)}{\partial A_n(i,j)}$.

\item [] (2.) The influence of a node $i$ in knowledge segment $KS_n$ can be calculated as $I(N_n(i)) = \sum_{k \in N(n)} -2(h_{e_{k,n}} - h_{c_{k,n}})\frac{\partial sim(KS_n, KS_k)}{\partial N_n(i)}$.

\item [] (3.) The influence of a node attribute $j$ in knowledge segment $KS_n$ can be calculated as $I(N_n^j(i,i)) = \sum_{k \in N(n)} -2(h_{e_{k,n}} - h_{c_{k,n}})\frac{\partial sim(KS_n, KS_k)}{\partial N_n^j(i,i)}$.

\end{lemma}

\textbf{Second}, after we find all the key elements, we check the consistency of the semantic matching subgraph according to these key elements. The steps are as follows. 
For each pair of knowledge segments of the  semantic matching subgraph, if their key elements overlapping rate is greater than a threshold (60\%), we check the consistency of this pair. Suppose the corresponding triples are <${\tt s_1}$, ${\tt p_1}$, ${\tt o_1}$> and <${\tt s_2}$, ${\tt p_2}$, ${\tt o_2}$>, respectively. We check if <${\tt s_1}$, {\tt isTypeOf}, ${\tt s_2}$> or <${\tt s_2}$, {\tt isTypeOf}, ${\tt s_1}$> is true. If both of them are false, we skip this pair clues because this clue pair does not belong to C3 or C4. Otherwise, we check if <${\tt o_1}$, {\tt isTypeOf}, ${\tt o_2}$> or <${\tt o_2}$, {\tt isTypeOf}, ${\tt o_1}$> is true. If both of them are false, we say this query graph has collective inconsistency. When checking the truthfulness of triples (e.g., <${\tt s_1}$, {\tt isTypeOf}, ${\tt s_2}$>, 
<${\tt s_2}$, {\tt isTypeOf}, ${\tt s_1}$>, 
<${\tt o_1}$, {\tt isTypeOf}, ${\tt o_2}$> and <${\tt o_2}$, {\tt isTypeOf}, ${\tt o_1}$>), 
we use the same method (i.e., transferred information amount in Eq.~
\eqref{eq:transinfo}) as in pairwise comparative reasoning. 

\vspace{-0.5\baselineskip}
\section{Experimental Results} \label{experiment}

In this section, we present the experimental evaluations. All the experiments are designed to answer the following two questions:

\begin{itemize}
	\item {\bf Q1. Effectiveness.} How effective are the proposed reasoning methods, including both point-wise methods (\gchecker\ basics) and comparative reasoning methods? 
	\item {\bf Q2. Efficiency.} How fast are the proposed methods? 
\end{itemize}

We use the Yago dataset ~\cite{yago}. \footnote{It is publicly available at \url{https://www.mpi-inf.mpg.de/departments/databases-and-information-systems/research/yago-naga}.}
It contains 12,430,705 triples, 4,295,825 entities and 39 predicates.
All the experiments are conducted on a moderate desktop with an Intel Core-i7 3.00GHz CPU and 64GB memory. The source code will be released upon publication of the paper. 

\vspace{-0.5\baselineskip}
\subsection{\gchecker\ Basics}

\hide{
\begin{table}[ht]
\caption{Top-{\em 10} predicates in Yago}
\vspace{-1\baselineskip}
\subfloat[Highest Entropy]
{\begin{tabular}{|c|c|}
\hline
Rank & Predicate \\
\hline
1 & edited   \\
2 & isConnectedTo    \\
3 & actedIn    \\
4 & playsFor    \\
5 & dealsWith    \\
6 & directed    \\
7 & exports    \\
8 & isAffiliatedTo    \\
9 & wroteMusicFor    \\
10 & hasNeighbor      \\
\hline
\end{tabular}}
\subfloat[Lowest Entropy]
{\begin{tabular}{|c|c|}
\hline
Rank & Predicate \\
\hline
1 & diedIn     \\
2 & hasGender      \\
3 & hasCurrency      \\
4 & wasBornIn      \\
5 & hasAcademicAdvisor      \\
6 & isMarriedTo      \\
7 & hasCapital      \\
8 & hasWebsite      \\
9 & isPoliticianOf      \\
10 & isCitizenOf        \\
\hline
\end{tabular}}
\label{predicate-entropy}
\vspace{-1\baselineskip}
\end{table}
}

We start with evaluating the effectiveness of the proposed predicate entropy. 
The top-{\em 10} predicates with the highest predicate entropy in Yago dataset are {\tt edited}, {\tt isConnectedTo}, {\tt actedIn}, {\tt playsFor}, {\tt dealsWith}, {\tt directed}, {\tt exports}, {\tt isAffiliatedTo}, {\tt wroteMusicFor} and {\tt hasNeighbor}. Predicates like {\tt actedIn}, {\tt playFor}, {\tt hasNeighbor} have a very high entropy. 
The reason is that these predicates not only occur commonly in the Yago knowledge graph, but also 
have a high degree of uncertainty.
It is consistent with our hypothesis that these predicates provide little semantic information about the entities around them.
On the contrary, The top-{\em 10} predicates with the lowest predicate entropy in Yago dataset are {\tt diedIn}, {\tt hasGender}, {\tt hasCurrency}, {\tt wasBornIn}, {\tt hasAcademicAdvisor},  {\tt isPoliticianOf}, {\tt isMarriedTo}, {\tt hasCaptal}, {\tt hasWebsite}, and {\tt isCitizenOf}. Predicates like {\tt diedIn}, {\tt wasBornIn}, {\tt isMarriedTo}, {\tt isPoliticianOf} have a very low entropy.
They provide specific and detailed background information about the entities around them.

\hide{
\begin{figure}[ht]
\vspace{-1.5\baselineskip}
  \subfloat[exports]{
  \vspace{-0.5\baselineskip}
	\begin{minipage}[c][0.5\width]{
	   0.2\textwidth}
	   \centering
	   \includegraphics[width=1\textwidth]{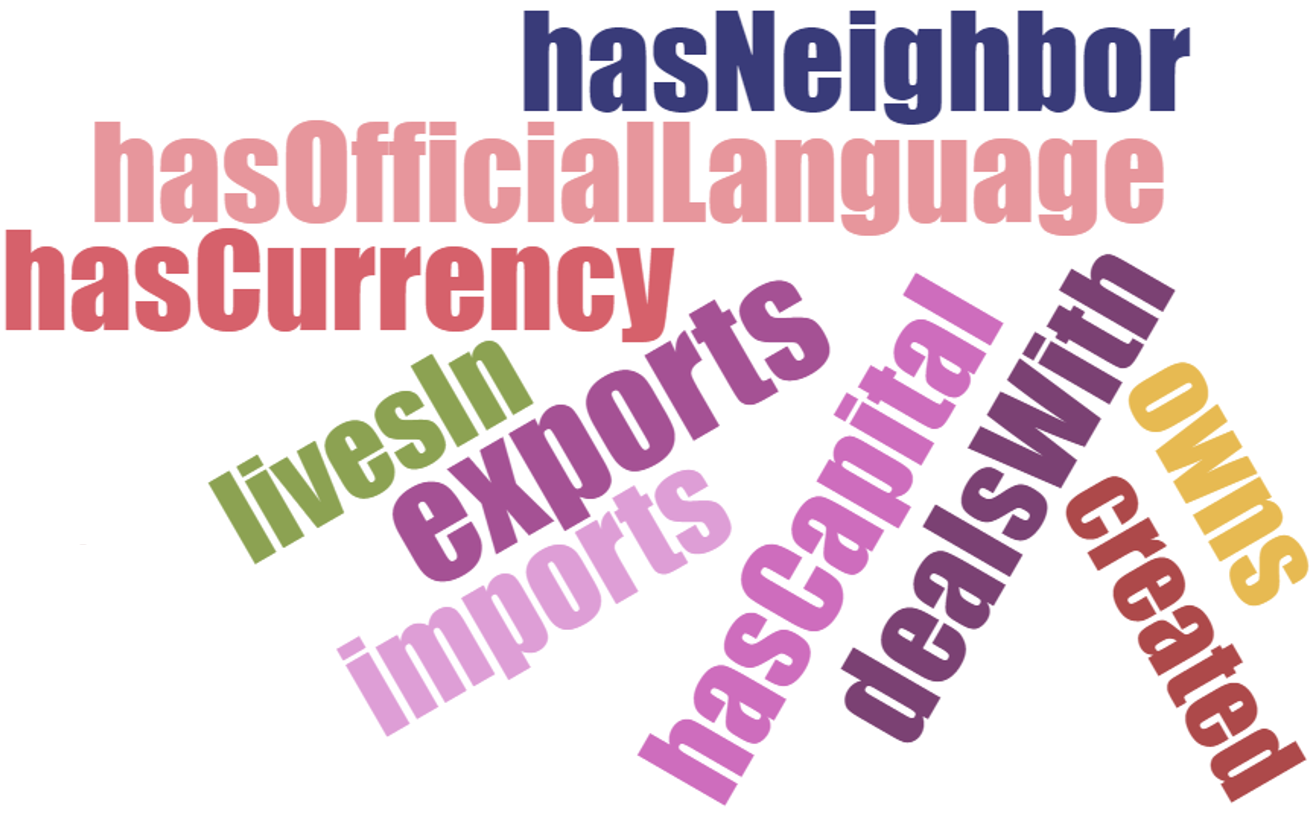}
	\end{minipage}}
 \hfill 	
  \subfloat[livesIn]{
  \vspace{-0.5\baselineskip}
	\begin{minipage}[c][0.5\width]{
	   0.2\textwidth}
	   \centering
	   \includegraphics[width=1\textwidth]{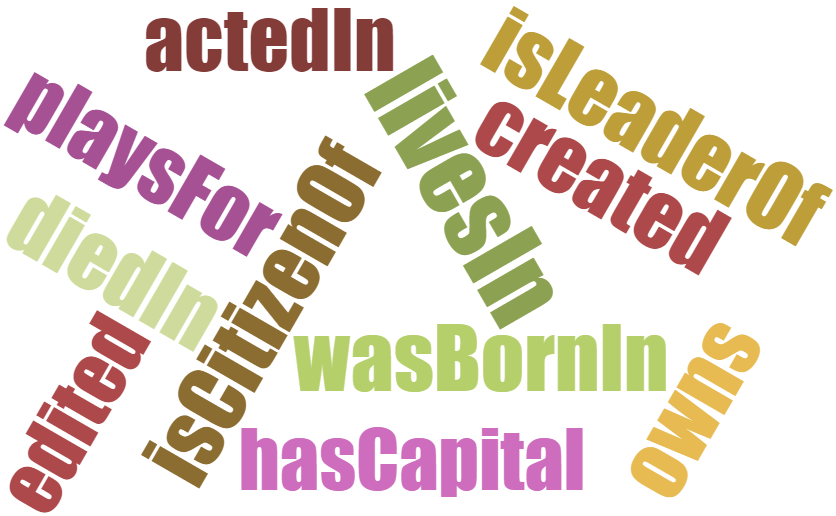}
	\end{minipage}}
\vspace{-1\baselineskip}
\caption{Top-{\em 10} most similar predicates in Yago. \hh{if we need space, we can also turn this figure into inline text}}
\label{word-cloud}
\vspace{-1\baselineskip}
\end{figure}
}

Next, we evaluate the proposed predicate-predicate similarity. 
The top similar predicates w.r.t. {\tt exports} by our method include {\tt imports}, {\tt hasOfficialLanguage}, {\tt dealsWith}, all of which have a high similarity with {\tt exports}. They all provide specific semantic information about {\tt exports}.
Likewise, the top similar predicates w.r.t. {\tt livesIn} include {\tt wasBornIn}, {\tt isCitizenOf}, {\tt diedIn}, all of which are closely related to {\tt livesIn}. 
These results show that the proposed TF-IDF based method can effectively measure the similarity between different predicates.

\begin{figure}
	\centering
	\includegraphics[width=0.4\textwidth, height=0.25\textwidth]{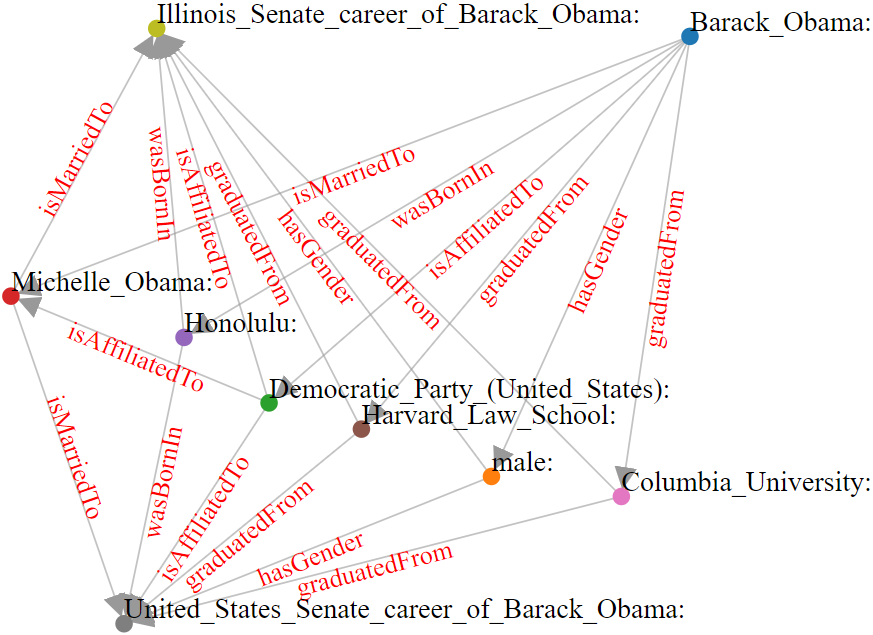}
	\vspace{-1\baselineskip}
	\caption{Node-specific knowledge segment of {\tt Barack Obama}.}
	\label{nibble}
	\vspace{-1\baselineskip}
\end{figure}

Figure ~\ref{nibble} shows the node-specific knowledge segment found by our method w.r.t. the query node {\tt Barack Obama}. We can see that the extracted knowledge segment provides critical semantics of the query node {\tt Barack Obama}. For example, {\tt Barack Obama} graduated from {\tt Harvard Law School} and {\tt Columbia University}; his wife is {\tt Michelle Obama}; he is affiliated to the democratic party; he was the senator of united states and he was born in {\tt Honolulu}. 
The result shows that node-specific knowledge segment is able to capture semantic context of the query node/entity. 

\subsection{Pair-wise Comparative Reasoning}\label{exp-pair-section}

Here, we evaluate the effectiveness of the proposed pair-wise comparative reasoning.
We first give an example to show how it works, then we evaluate it on several subsets of Yago.
Consider a fake news which says {\em ``The white house will participate in the operation mountain thrust, because the white house wants to punish the iraqi army."} From this news, we can extract two query clues, including <{\tt White House}, {\tt participatedIn}, {\tt Operation Mountain Thrust}> and <{\tt White House}, {\tt punish}, {\tt Iraqi Army}>.
Figure ~\ref{iraq_inconsistency-example} shows these two corresponding knowledge segments. Table ~\ref{iraq_pair-node-attr} shows the node attribute influence value of $KS_1$ and $KS_2$, respectively \footnote{We skip the node influence value and the edge influence value due to space limit.}.
We can see from Table ~\ref{iraq_pair-node-attr} that
for $KS_1$, the top-{\em 50\%} elements with the highest node attribute influence
are {\tt Washington,D.C}, {\tt United States President}, {\tt White House} and {\tt United States}.
For $KS_2$, the top-{\em 50\%} elements with the highest node attribute influence are {\tt White House}, {\tt Washington,D.C}, and {\tt United States}.
Because all the import elements with the highest node attribute influence of $KS_2$ also belong to $KS_1$, the key elements overlapping rate for node attribute is 100\%.
For the top-{\em 50\%} elements with the highest node influence, we obtain the same result.
As for the top-{\em 50\%} edges of $KS_1$ with the highest influence, there is one edge (<{\tt United States}, {\tt hasCapital}, {\tt Washington,D.C}>) which also belongs to the top-50\% edges of $KS_2$
\hide{
\footnote{We skip the details due to space limit. An extended version with additional experiments as well as a video demo can be found at
\url{https://github.com/lihuiliullh/KompaRe}
}
}
.
Therefore, the key elements overlapping rate of $KS_1$ and $KS_2$ is $\frac{1 + 1 + \frac{1}{3}}{3} = \frac{7}{9} > 60\%$. This means that these two clues refer to the same thing. 

\begin{table}[]
\vspace{-1\baselineskip}
\caption{Pairwise node attribute influence ranking}
\vspace{-1\baselineskip}
\scriptsize
\subfloat[Node attribute influence: $KS_1$]
{\begin{tabular}{|c|c|c|}
\hline
 Rank & Predicate & value \\
\hline
1 & Washington, D.C & 4.49 $e^{-5}$\\
2 & United States President &  3.92 $e^{-5}$  \\
3 & White House & 3.90$e^{-5}$   \\
4 & United States & 3.87$e^{-5}$    \\
5 & United States Army  & 1.96$e^{-5}$    \\
6 & Afghanistan & 1.95$e^{-5}$    \\
7 & Operation Mountain Thrust   &  1.95$e^{-5}$  \\
\hline
\end{tabular}}
\subfloat[Node attribute influence: $KS_2$]{
\renewcommand\arraystretch{1.26}%
\begin{tabular}{|c|c|c|}
\hline
Rank & Predicate & value \\
\hline
1 & Washington, D.C & 4.51$e^{-5}$    \\
2 & White House   &  3.90$e^{-5}$  \\
3 & United States  & 3.87$e^{-5}$    \\
4 & United States President &  3.83$e^{-5}$  \\
5 & Iraq & 1.97$e^{-5}$    \\
6 & Iraqi Army & 1.86$e^{-5}$   \\
\hline
\end{tabular}}
\label{iraq_pair-node-attr}
\vspace{-2\baselineskip}
\end{table}

\begin{figure*}[h]
\hspace*{\fill}%
\begin{minipage}[t]{0.4\textwidth}
\centering
\vspace{0pt}
\includegraphics[width=1.2\textwidth,height=0.64\textwidth]{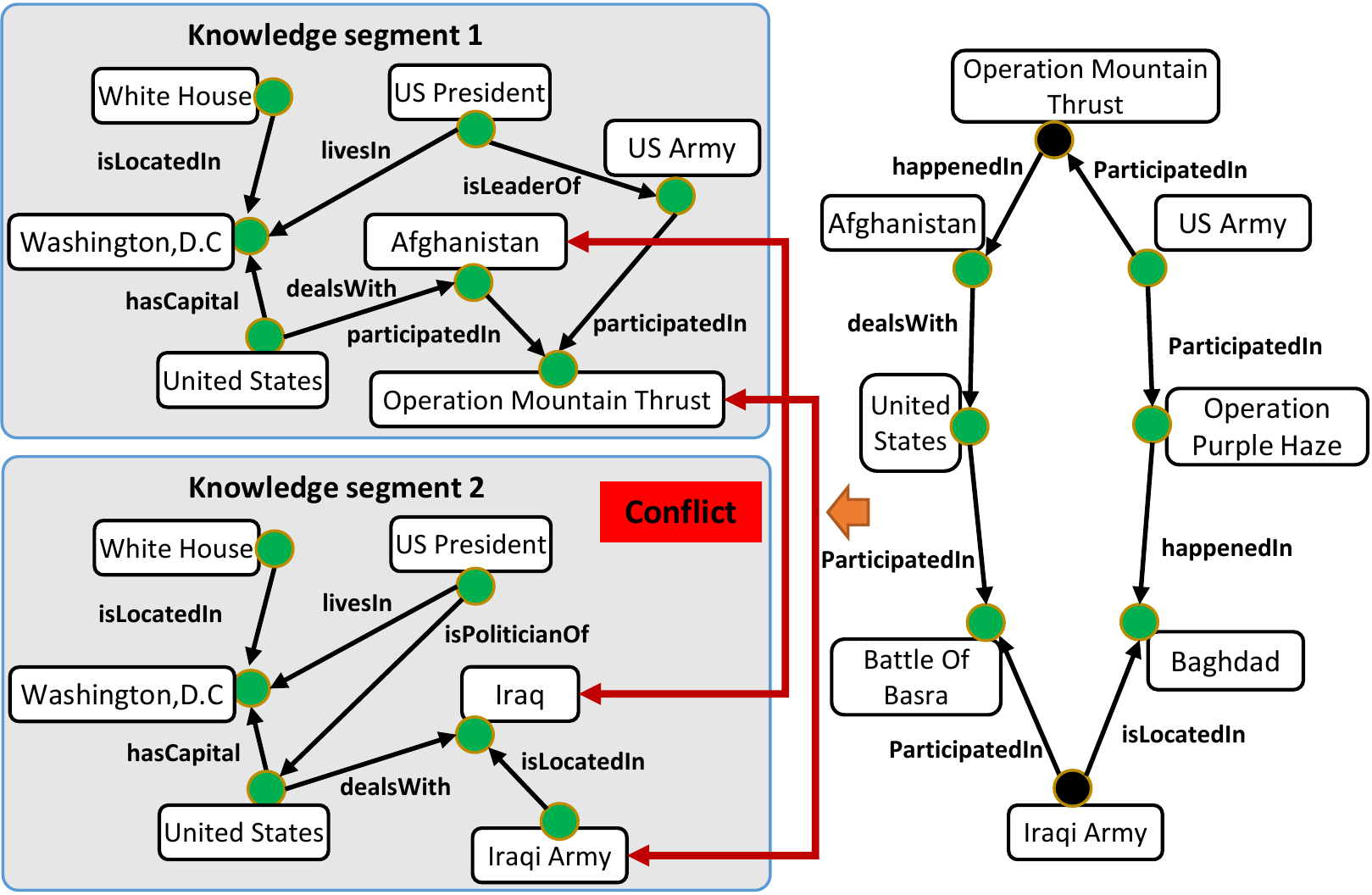}
\vspace{-2\baselineskip}
\caption{Pair-wise comparative reasoning results.}
\label{iraq_inconsistency-example}
\end{minipage}%
\hspace{20.00mm}
\begin{minipage}[t]{0.425\textwidth}
\centering
\vspace{0pt}
\includegraphics[width=1.1\textwidth,height=0.6\textwidth]{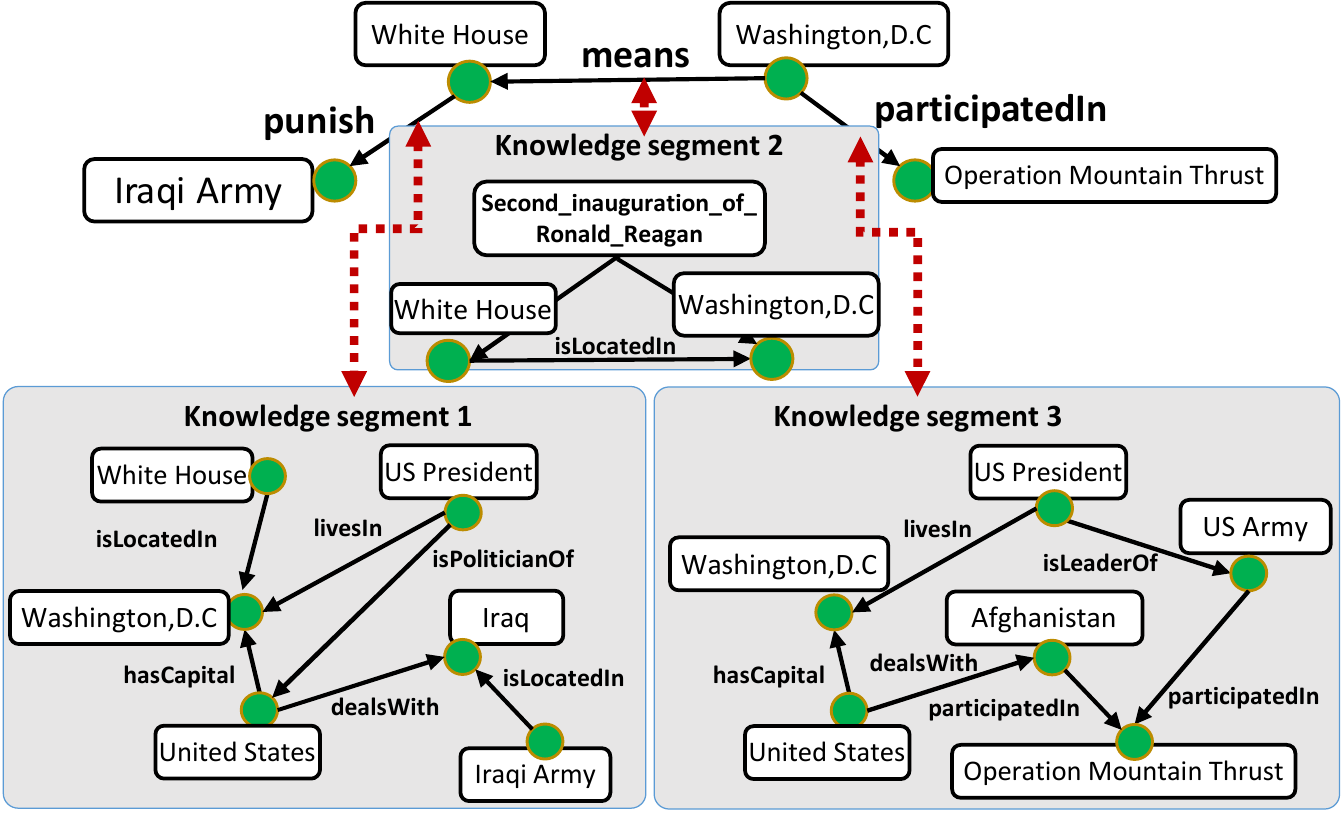}
\vspace{-2\baselineskip}
\caption{Collective comparative reasoning results.}
\label{iraq_exp-coll}
\end{minipage}%
\hspace*{\fill}
\vspace{-1\baselineskip}
\end{figure*}

We further check if there is any inconsistency between them. To this end, we extract the knowledge segment for <{\tt Operation Mountain Thrust}, {\tt isTypeOf}, {\tt Iraqi Army}> and <{\tt Iraqi Army}, {\tt isTypeOf}, {\tt Operation Mountain Thrust}>. 
The right part of Figure ~\ref{iraq_inconsistency-example} shows the knowledge segment for <{\tt Operation Mountain Thrust}, {\tt isTypeOf}, {\tt Iraqi Army}>. We obtain the same knowledge segment for <{\tt Iraqi Army}, {\tt isTypeOf}, {\tt Operation Mountain Thrust}>. 
The proposed TF-IDF predicate-predicate similarities between {\tt isTypeOf} and {\tt happendIn}, {\tt dealsWith}, {\tt participatedIn}, {\tt isLocatedIn} are 0.767, 0.697, 0.869 and 0.870, respectively.
Based on that, we have \textrm{infTrans}({\tt Operation Mountain Thrust}, {\tt Iraqi Army}) = \textrm{infTrans}({\tt Iraqi Army}, {\tt Operation Mountain Thrust}) = $ 0.505 < 0.700$. This means that "{\tt Operation Mountain Thrust}" and "{\tt Iraqi Army}" are two different things. 
We obtain the same result for triple <{\tt Iraqi Army}, {\tt isTypeOf}, {\tt Operation Mountain Thrust}>. Therefore, we conclude that 
the two given clues are inconsistent.

\begin{table}
	\centering
	\caption{Accuracy of pair-wise comparative reasoning.}
	\vspace{-1\baselineskip}
	\fontsize{8}{9}\selectfont
	\begin{tabular}{|c|c|c|c|c|c|c|c|c|c|c|c|c|}
	\hline
	Yago Dataset          & \multicolumn{2}{|c|}{Positive}   & \multicolumn{2}{|c|}{Negative}      \\ \hline
	Query Category      & \# of Queries & Accuracy & \# of Queries & Accuracy \\ \hline
	Birth Place     & 452 &  0.906 &  89 & 0.785  \\ \hline
	Graduated College & 386 & 0.839 & 68 & 0.929  \\ \hline
    Family Members       & 356 & 0.912  & 69 & 0.824  \\ \hline
    Live Place      & 447 & 0.871  & 79 & 0.90  \\ \hline
    Citizenship     & 389 & 0.841 & 64 & 0.941  \\ \hline
	\end{tabular}
\label{pair_dataset}
\end{table}

\hide{
\begin{table}
	\centering
	\caption{Precision of pair-wise comparative reasoning.}
	\fontsize{8}{9}\selectfont
	\begin{tabular}{|c|c|c|c|c|c|c|c|c|c|c|c|c|}
	\hline
	Yago Dataset          & \multicolumn{2}{|c|}{Positive}   & \multicolumn{2}{|c|}{Negative}      \\ \hline
	Pairwise Type          & \# of Queries & Accuracy & Size & Accuracy \\ \hline
	(graduatedFrom, isCitizenOf) & 386 & 0.839 & 68 & 0.929  \\ \hline
    (hasChild, isMarriedTo)  & 356 & 0.912  & 69 & 0.824  \\ \hline
    (livesIn, isCitizenOf)  & 447 & 0.871  & 79 & 0.90  \\ \hline
    (wasBornIn, isCitizenOf) & 389 & 0.841 & 64 & 0.941  \\ \hline
    (wasBornIn, livesIn) & 452 &  0.906 &  89 & 0.785  \\ \hline
	\end{tabular}
\label{pair_dataset}
\end{table}
}

We test Pair-wise comparative reasoning method on 5 query sets. Table ~\ref{pair_dataset} gives the details of the results. 
For each positive query set, it contains a set of queries which describe the truth, while for each negative query set, it contains a set of queries which describe the false fact. For example, in query set "Birth Place", <{\tt Barack Obama}, {\tt wasBornIn}, {\tt Honolulu}> and <{\tt Barack Obama}, {\tt wasBornIn}, {\tt United States}> is a positive query pair, while <{\tt Barack Obama}, {\tt wasBornIn}, {\tt Honolulu}> and <{\tt Barack Obama}, {\tt wasBornIn}, {\tt Canada}> is an negative query pair.
The accuracy is defined as $\frac{N}{M}$ where $N$ is the number of queries correctly classified by pair-wise comparative reasoning
and $M$ is the total number of queries. As we can see, the average accuracy of pair-wise comparative reasoning is more than 85\%.

\hide{
\begin{figure}[]
	\centering
	\includegraphics[width=0.47\textwidth]{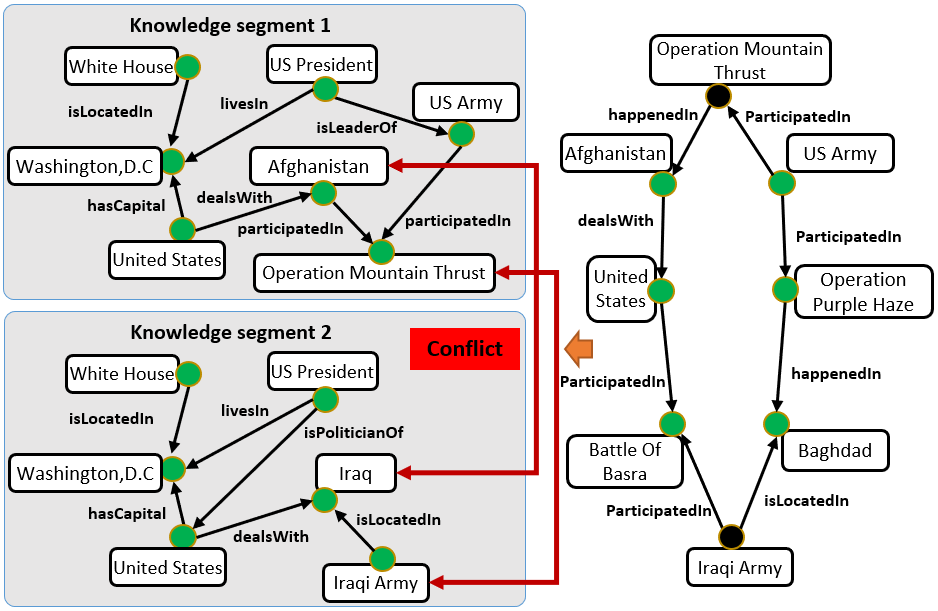}
	\caption{Pair-wise comparative reasoning results.}
	\label{iraq_inconsistency-example}
	\vspace{-1\baselineskip}
\end{figure}

\begin{figure}[]
	\centering
	\includegraphics[width=0.47\textwidth]{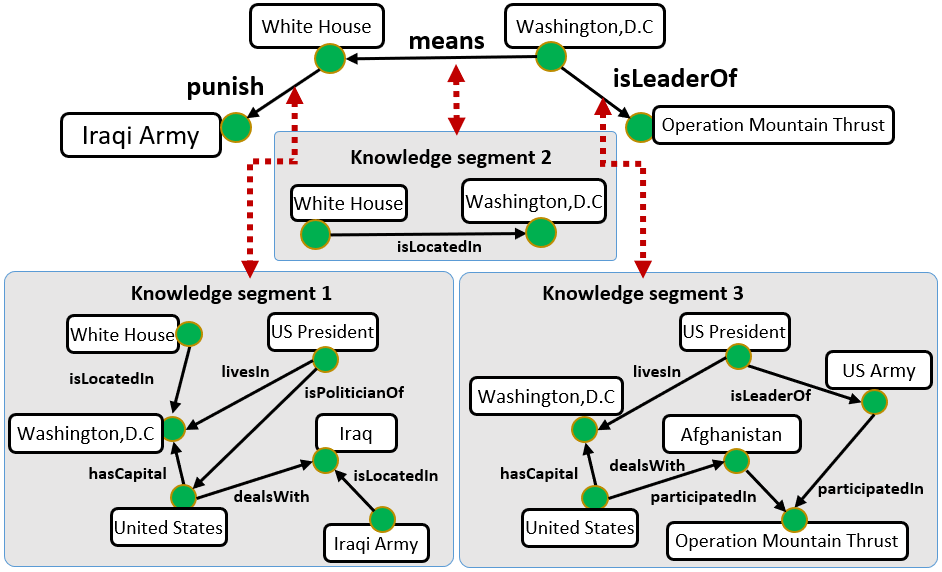}
	\caption{Collective comparative reasoning results.}
	\label{iraq_exp-coll}
	\vspace{-1\baselineskip}
\end{figure}
}

\vspace{-0.6\baselineskip}
\subsection{Collective Comparative Reasoning}\label{exp-coll-section}

Here, we evaluate the effectiveness of the proposed collective comparative reasoning. 
We first give an example to show how it works, then we evaluate it on several subsets of Yago.
We test a query graph with three edges, including <{\tt White House}, {\tt punish}, {\tt Iraqi Army}>, <{\tt Washington,D.C}, {\tt means}, {\tt White House}> and <{\tt Washington,D.C}, {\tt participatedIn}, {\tt Operation Mountain Thrust}>. 
Figure ~\ref{iraq_exp-coll} shows the query graph and the corresponding semantic matching subgraph. 
As we can see, if we use the pair-wise comparative reasoning method to check each pair of them, all of them are true, because none of them belong to C3 or C4. 
However, if we use the collective comparative reasoning method, we could detect the inconsistency in the query graph as follows.

If we check each pair of clues in the query graph, we find that the key elements overlapping rate between $KS_1$ and $KS_3$ is more than 60\%.
This is because the overlapping rates  are $66.6\%$ for node attribute influence, $100\%$ for node influence and $66.6\%$ for edge influence, which give the average overlapping rate $\frac{\frac{2}{3} + 1 + \frac{2}{3}}{3} > 60\%$. 

Based on this, we future check <{\tt Washington,D.C}, {\tt isTypeOf}, {\tt White House}> or <{\tt White House}, {\tt isTypeOf}, {\tt Washington,D.C}>. 
Our TF-IDF based predicate-predicate similarity between "{\tt isTypeOf}" and "{\tt isLocatedIn}" is 0.870. Thus, we have \textrm{infTrans}({\tt Washington,D.C}, {\tt White House}) $= 0.870 > 0.700$. This means that these two knowledge segments have the same subject. Finally, we check  <{\tt Operation Mountain Thrust}, {\tt isTypeOf}, {\tt Iraqi Army}> or <{\tt Iraqi Army}, {\tt isTypeOf}, {\tt Operation Mountain Thrust}>. According to the results in the previous subsection, we have that {\tt Iraqi Army} and {\tt Operation Mountain Thrust} are two different things. Therefore, we conclude that this query graph is inconsistent.

\begin{table}
	\centering
	\caption{Accuracy of collective comparative reasoning.}
	\vspace{-1\baselineskip}
	\fontsize{8}{9}\selectfont
	\begin{tabular}{|c|c|c|c|c|c|c|c|c|c|c|c|c|}
	\hline
	Yago Dataset          & \multicolumn{2}{|c|}{Positive}   & \multicolumn{2}{|c|}{Negative}      \\ \hline
	Query Category      & \# of Queries & Accuracy & \# of Queries & Accuracy \\ \hline
	Birth Place & 375 & 0.838 & 52 & 0.902  \\ \hline
	Citizenship  & 486 & 0.798  & 92 & 0.861  \\ \hline
    Graduated College & 497 & 0.711 & 56 & 0.882  \\ \hline
	\end{tabular}
\label{coll_dataset}
\vspace{-1\baselineskip}
\end{table}

We test collective comparative reasoning method on 3 query sets. Table ~\ref{coll_dataset} gives the details of the results. 
Different from the queries of pair-wise comparative reasoning which only contain two edges, each query of collective comparative reasoning contains 3 edges. 
For example, in query set "Birth Place", <{\tt Barack Obama}, {\tt wasBornIn}, {\tt Honolulu}>, <{\tt Barack Obama}, {\tt means}, {\tt United States Senate Barack Obama}> and <{\tt United States Senate Barack Obama}, {\tt wasBornIn}, {\tt United States}> is a positive query triad, while <{\tt Barack Obama}, {\tt wasBornIn}, {\tt Honolulu}>, <{\tt Barack Obama}, {\tt means}, {\tt United States Senate Barack Obama}> and <{\tt United States Senate Barack Obama}, {\tt wasBornIn}, {\tt Canada}> is an negative query triad.
The definition of the accuracy is the same as the previous section.
As we can see, the average accuracy of collective comparative reasoning is more than 82\%.

\hide{
\begin{table}
	\centering
	\caption{Average number of answer entities of test queries}
	\fontsize{8}{9}\selectfont
	\begin{tabular}{|c|c|c|c|c|c|c|c|c|c|c|c|c|}
	\hline
	Yago Dataset          & \multicolumn{2}{|c|}{Positive}   & \multicolumn{2}{|c|}{Negative}      \\ \hline
	Collective Type          & Size & Precision & Size & Precision \\ \hline
	(livesIn,  isCitizenOf)  & 486 & 0.798  & 92 & 0.861  \\ \hline
	(wasBornIn, isCitizenOf) & 375 & 0.838 & 52 & 0.902  \\ \hline
    (graduatedFrom, isCitizenOf) & 497 & 0.711 & 56 & 0.882  \\ \hline
	\end{tabular}
\label{coll_dataset}
\end{table}
}

\subsection{\gchecker\ Efficiency}

The runtime of knowledge segment extraction depends on the size of the underlying knowledge graphs. Among the three types of knowledge segments ({\tt f1}, {\tt f2} and {\tt f3} in Table~\ref{table:fun:summary}), subgraph-specific knowledge segment ({\tt f3}) is most time-consuming. Figure ~\ref{fig:runtime}(a) shows that its runtime scales near-linearly w.r.t. the number of nodes in the knowledge graph. Figure ~\ref{fig:runtime}(b) shows the runtime of comparative reasoning, where `Pair-wise' refers to the pairwise comparative reasoning, and the remaining bars are for collective comparative reasoning with $3$, $4$ and $5$ edges in the query graphs respectively. Notice that the runtime of comparative reasoning only depends on the size of the the corresponding knowledge segments which typically have a few or a few tens of nodes. In other words, the runtime of comparative reasoning is {\em independent} of the knowledge graph size. If the query has been searched before, the runtime is less than 0.5 second. \footnote{The system was deployed in May 2020.}

\begin{figure}\label{fig:runtime}
\vspace{-0.8\baselineskip}
\centering
\hide{
\subfloat[][Node-specific Knowledge Segment Extraction Runtime]{%
\label{node-runtime}%
\includegraphics[height=1.7in]{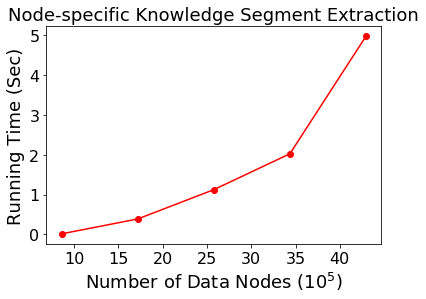}}%
\hspace{50pt}%
\subfloat[][Edge-specific Knowledge Segment Extraction Runtime]{%
\label{edge-runtime}%
\includegraphics[height=1.7in]{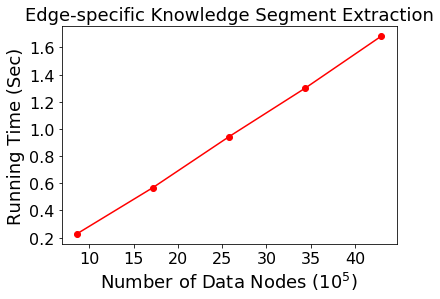}}%
}
\subfloat[h][Subgraph-specific KS extraction]{%
\label{semantic-runtime-data}%
\includegraphics[height=1.2in]{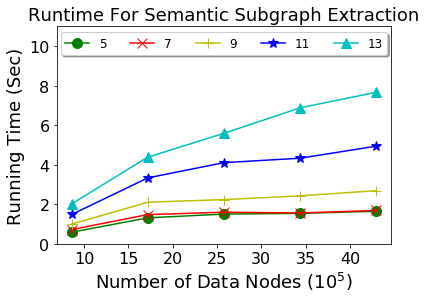}}%
\subfloat[][Comparative reasoning]{%
\label{comp-time}%
\includegraphics[height=1.2in]{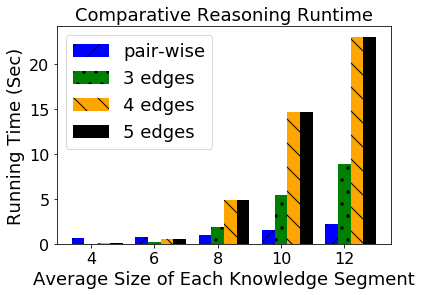}}

\vspace{-1\baselineskip}
\caption[A set of four subfigures.]{
Runtime of \gchecker\
}
\label{precision-4}%
\vspace{-2\baselineskip}
\label{fig:runtime}
\end{figure}

\vspace{-0.5\baselineskip}

\vspace{-0.3\baselineskip}
\section{Conclusions}

In this paper, we present a prototype system (\gchecker) for knowledge graph comparative reasoning. 
\gchecker\ aims to complement and expand the existing point-wise reasoning over knowledge graphs by inferring commonalities and inconsistencies of multiple pieces of clues. 
The developed prototype system consists of three major components, including its UI, online reasoning and offline mining. At the heart of the proposed \gchecker\ are a suite of core algorithms, including predicate entropy, predicate-predicate similarity and semantic subgraph matching for knowledge segment extraction; and influence function, commonality rate, transferred information amount for both pairwise reasoning and collective reasoning. 
The experimental results demonstrate that the developed \gchecker\ (1) can effectively detect semantic inconsistency, and (2) scales near linearly with respect to the knowledge graph size. 

\hide{
\begin{table}[ht]
\caption{Collective Node Attribute Influence Ranking}
\small
\subfloat[Highest absolute Influence in KS1]
{\begin{tabular}{|c|c|c|}
\hline
Ranking & Predicate & value \\
\hline
1 & 600M & -0.00018  \\
2 & Air Force &  -0.00019  \\
3 & Helicopter & -0.00021    \\
4 & US Army & -0.00021    \\
5 & President      & -0.00035    \\

\hline
\end{tabular}}
\subfloat[Highest absolute Influence in KS2]
{\begin{tabular}{|c|c|c|}
\hline
Ranking & Predicate & value \\
\hline
1 & Barack Obama   &   -0.00048 \\
2 & President      &   -0.00065 \\
\hline
\end{tabular}}
\subfloat[Highest absolute Influence in KS3]
{\begin{tabular}{|c|c|c|}
\hline
Ranking & Predicate & value \\
\hline
1 & 6000M & -0.00015  \\
2 & Air Force &  -0.00017  \\
3 & Air Force One & -0.00018    \\
4 & US Army & -0.00018    \\
5 &  Barack Obama & -0.00031 \\
6 & President     & -0.00032    \\
\hline
\end{tabular}}
\label{coll-node-attr}
\end{table}
}

\vspace{-0.7\baselineskip}
\bibliographystyle{ACM-Reference-Format}
\bibliography{008reference.bib}

\clearpage

\end{document}